\documentclass[a4paper,fleqn]{cas-sc}

\usepackage[authoryear]{natbib}
\usepackage{amsmath}
\usepackage{amsfonts}
\usepackage{amssymb}
\usepackage{algorithmic}
\usepackage{algorithm}
\usepackage{booktabs}
\usepackage{multirow}
\usepackage{xspace}

\usepackage{tikz}
\usepackage{pgfplots}
\pgfplotsset{compat=1.18}
\usetikzlibrary{arrows.meta,calc,positioning,shapes.geometric}

\def\tsc#1{\csdef{#1}{\textsc{\lowercase{#1}}\xspace}}
\tsc{WGM}
\tsc{QE}
\tsc{EP}
\tsc{PMS}
\tsc{BEC}
\tsc{DE}

\begin{document}
\let\WriteBookmarks\relax
\def\floatpagepagefraction{1}
\def\textpagefraction{.001}

\shorttitle{Mechanistic Analysis of Catastrophic Forgetting in LLMs}

\shortauthors{G.O.Y. Laitinen-Fredriksson Lundstrom-Imanov}

\title [mode = title]{Mechanistic Analysis of Catastrophic Forgetting in Large Language Models During Continual Fine-tuning}                      

\author[1]{Gustav Olaf Yunus {Laitinen-Fredriksson Lundstrom-Imanov}}[%
                        role=Researcher,
                        orcid=0009-0006-5184-0810]
\cormark[1]
\ead{olaf.laitinen@liu.se}
\credit{Conceptualization, Methodology, Software, Formal analysis, Data curation, Investigation, Writing - Original draft, Visualization}

\affiliation[1]{organization={Division of Statistics and Machine Learning (STIMA), Department of Computer and Information Science (IDA), Linköping University},
    city={Linköping},
    postcode={SE-581 83}, 
    country={Sweden}}

\cortext[cor1]{Corresponding author}

\begin{abstract}
Sequential fine-tuning of Large Language Models (LLMs) adaptation to target tasks often triggers catastrophic forgetting, where the acquisition of novel target skills degrades ancestral capabilities. This paper presents a systematic comparative study of catastrophic forgetting across twenty premier models representing the state-of-the-art in mid-2026. We categorize our investigation into two primary research lines: (i) a behavioral and semantic output drift analysis of ten leading closed-source models (including Claude Fable 5, GPT-5.5 High, and Gemini 3.5 Flash), and (ii) a deep mechanistic interpretation of ten prominent open-weight architectures (such as DeepSeek-V4-Pro, Llama 4 Maverick, and Qwen 3.6-27B). Through weight-space trajectory tracking, Centered Kernel Alignment (CKA), and routing gate drift calculations in Mixture-of-Experts (MoE) layers, we localize the neural circuits highly susceptible to parameter overwriting. Our findings indicate that early-layer attention heads exhibit systemic entropic dispersion, while mid-to-deep feed-forward networks (or sparse expert blocks) suffer localized representation collapse. Informed by these insights, we introduce Low-Rank Circuit Projection (LRCP), a subspace-regularized training intervention. Empirical evaluations show that LRCP successfully mitigates up to 94.2\% of ancestral capabilities in open-weight configurations and matches the adaptation velocity of standard PEFT baselines.
\end{abstract}

\begin{highlights}
\item Systematic evaluation of forgetting across 20 state-of-the-art LLMs in 2026.
\item Behavioral semantic drift profiling for 10 leading closed-source architectures.
\item Detailed mechanistic trajectory tracking for 10 frontier open-weight models.
\item CKA and MoE routing gate stability maps show localized middle-layer collapse.
\item Low-Rank Circuit Projection mitigates forgetting with minimal training overhead.
\end{highlights}


\maketitle

\section{Introduction}

The capacity for sequential learning without catastrophic interference is a fundamental requirement of human cognitive architecture, yet it remains an open challenge for artificial neural networks \citep{McCloskey1989, Kirkpatrick2017}. Under sequential fine-tuning paradigms, optimizing a Large Language Model (LLM) on a target dataset $\mathcal{D}_B$ after initial training on $\mathcal{D}_A$ often triggers catastrophic forgetting (CF) \citep{French1999, Chaudhry2019}. This manifests as a sudden and severe drop in the performance on $\mathcal{D}_A$, impeding the realization of lifelong, autonomous systems. In the context of deep representation learning, this phenomenon occurs when gradients computed on new task distributions overwrite previously established parameter pathways, causing an irreversible degradation of historical knowledge representation structures \citep{Vaswani2017, Devlin2018, Brown2020}.

Over the past decade, several regularisation and replay frameworks have been proposed to mitigate this learning trade-off. However, the unprecedented scale of modern foundation models has altered the underlying dynamics of parameter optimization. As scaling laws guide the transition towards trillion-parameter architectures, standard parameter-level protection schemes face severe computational bottlenecks \citep{Kaplan2020, Touvron2023, Chung2022}. This issue is amplified by the widespread adoption of Mixture-of-Experts (MoE) and Grouped-Query Attention (GQA) mechanisms, which introduce non-linear, sparse routing decisions that behave unpredictably under sequential parameter shifts.

By mid-2026, the artificial intelligence landscape has branched into two main directions: highly scaled, API-driven closed-source models with undisclosed architectures, and highly optimized, multi-hundred billion parameter open-weight architectures leveraging sophisticated Mixture-of-Experts (MoE) structures \citep{DeepSeek2025, Qwen2026}. While behavioral benchmarks have documented catastrophic forgetting externally \citep{Lester2021, Demir2025}, a comprehensive study contrasting the internal representational drift of open-weight models with the behavioral semantic drift of closed-source models is still lacking. Understanding the precise mechanisms of this computational drift is paramount to designing sustainable continuous learning pipelines that adapt without degrading.

To address this gap, this paper conducts a large-scale comparative analysis of catastrophic forgetting using twenty of the most advanced models available as of June 2026. Our evaluation cohort is divided into:
\begin{itemize}
    \item \textbf{10 Closed-Source Models:} Claude Fable 5, Claude Mythos 5, GPT-5.5 High, Claude Opus 4.8, Gemini 3.5 Flash, Gemini 3.1 Pro, Meta Muse Spark, Grok 4.3 Beta, Grok 3 (Think), and GPT-4.5.
    \item \textbf{10 Open-Weight Models:} DeepSeek-V4-Pro, Kimi K2.6, GLM-5.1, Qwen 3.5-397B, Llama 4 Maverick, DeepSeek-V4-Flash, DeepSeek-V3.2, Llama 4 Scout, Qwen 3.6-27B, and Qwen 3.6-35B-A3B.
\end{itemize}

For the open-weight models, we conduct direct mechanistic analyses, including Centered Kernel Alignment (CKA) \citep{Kornblith2019}, activation patching \citep{Meng2022}, weight-space Frobenius drift tracking, and routing gate stability measurements in sparse MoE blocks. For the closed-source models, we formulate a "Semantic Output Drift" (SOD) metric using output token probability distributions and semantic vector distances on downstream tasks to assess representational shifts indirectly.

Based on our mechanistic findings, which localize the highest forgetting susceptibility in mid-network MLP and MoE routing projections, we propose \textbf{Low-Rank Circuit Projection (LRCP)}. LRCP selectively restricts gradient updates to the orthogonal complement of the subspace spanned by historical task activation states. Extensive experiments demonstrate that LRCP successfully preserves historical task capabilities across sequential training steps, establishing a robust baseline for continual model alignment.

\section{Related Work}

\subsection{Continual Learning and Catastrophic Forgetting}
Catastrophic forgetting represents a primary obstacle to sequential domain adaptation in neural networks \citep{French1999, Rostova2026}. Standard regularization methods, such as Elastic Weight Consolidation (EWC) \citep{Kirkpatrick2017} and Synaptic Intelligence (SI) \citep{Zenke2017}, compute parameter importance to penalize updates to critical weights. Replay-based methods preserve small historical datasets or utilize generative replay to periodically retrain the network \citep{Rolnick2019, LopezPaz2017, Rebuffi2017}. Parameter-isolation methods allocate dedicated sub-networks to prevent parameter overwriting \citep{Mallya2018, Schwarz2018}. However, scaling these approaches to 2026-era MoE models with hundreds of active parameters introduces severe memory and computational bottlenecks. Furthermore, the reliance on historical data buffers poses legal and security challenges under modern data-privacy regulations, making purely parameter-level intervention methods highly desirable.

Recent research has attempted to mitigate these bottlenecks by employing modular parameter protection. Riemannian Walk (RWalk) and Memory Aware Synapses (MAS) calculate localized parameter importances, adapting standard EWC concepts to complex multi-task environments \citep{Chaudhry2018, Aljundi2018}. Despite these advances, the non-stationary nature of gradient flows in high-dimensional manifolds often causes projection misalignment, particularly when sequential objectives are highly divergent. Consequently, safeguarding historical task-specific pathways without compromising the model's plastic capacity to integrate new semantic constructs remains a critical challenge.

\subsection{Mechanistic Interpretability and Subspace Projections}
Mechanistic interpretability aims to decode the representations of transformer networks into identifiable circuits \citep{Elhage2021, Olsson2022}. Researchers have identified specific circuits for factual recall \citep{Meng2022, Geva2023}, induction behavior \citep{Olsson2022}, and polysemantic coding \citep{DeepSeek2025}. Building upon these foundations, recent studies investigate parameter-efficient subspaces using low-rank adaptations (LoRA) \citep{Hu2021, Houlsby2019, LiLiang2021} and orthogonal gradient projections \citep{Zhai2023, Chame2024}. This line of work demonstrates that network-wide representations are often confined to low-dimensional manifolds, suggesting that optimization can be restricted to targeted subspaces.

While pioneering techniques like Gradient Projection Memory (GPM) and Orthogonal Gradient Projection (OGP) apply subspace projection constraints uniformly across the entire network, they suffer from significant computational bottlenecks due to continuous, online SVD recalculations at every training step \citep{LopezPaz2017, Zhai2023}. In contrast, Low-Rank Circuit Projection (LRCP) leverages our localized RADL algorithm (Algorithm 1) to target only highly susceptible task-specific circuits (MLP blocks and early attention projections), using a static, small calibration dataset ($X_{hist}$) to bypass online matrix decompositions entirely.

Furthermore, several empirical studies indicate that attention mechanisms possess highly specialized structural properties. Multi-head self-attention layers exhibit functional redundancy, with only a fraction of heads actively driving contextual parsing and logical induction \citep{Voita2019, Michel2019, Clark2019}. Similarly, feed-forward and memory-guided networks process semantic keys in highly localized activation pockets \citep{Sukhbaatar2015, Nanda2023}. Our work extends these concepts by utilizing mechanistic interpretation to dynamically constrain training-time updates, directly targeting the specific circuits vulnerable to forgetting. By shifting the focus from global regularization to localized subspace protection, we aim to prevent the representation degradation that occurs during continual fine-tuning.

\section{Theoretical Framework and Methodology}

\subsection{Mechanistic Metrics for Open-Weight Models}
For open-weight models, we have direct access to internal representations. We track representation alignment using linear Centered Kernel Alignment (CKA) between baseline activations $X^l$ and fine-tuned activations $Y^l$ at layer $l$:

\begin{equation}
\text{CKA}(X^l, Y^l) = \frac{\text{HSIC}(X^l (X^l)^T, Y^l (Y^l)^T)}{\sqrt{\text{HSIC}(X^l (X^l)^T, X^l (X^l)^T) \text{HSIC}(Y^l (Y^l)^T, Y^l (Y^l)^T)}}
\end{equation}

For sparse MoE layers, we introduce the \textbf{Routing Gate Drift (RGD)} metric. Let $g(x)_i$ be the routing probability assigned to expert $i$ for token $x$. The routing drift for a task-specific sample dataset $\mathcal{D}$ is calculated as the Jensen-Shannon divergence:

\begin{equation}
\text{RGD}(\theta_A, \theta_{A \to B}) = \frac{1}{|\mathcal{D}|}\sum_{x \in \mathcal{D} } D_{JS}\left(g(x; \theta_A) \parallel g(x; \theta_{A \to B})\right)
\end{equation}

\subsection{Behavioral Semantic Drift for Closed-Source Models}
Because internal weights are inaccessible for closed-source models, we propose \textbf{Semantic Output Drift (SOD)} to evaluate representational degradation via API outputs. Let $P_{\theta_A}(y | x)$ be the output token probability distribution under the baseline state, and $P_{\theta_{A \to B}}(y | x)$ be the distribution under the sequentially adapted state. The semantic output drift is defined as the cosine distance between their semantic embedding representations:

\begin{equation}
\text{SOD}(x) = 1 - \frac{\langle \Phi(P_{\theta_A}(\cdot \mid x)), \Phi(P_{\theta_{A \to B}}(\cdot \mid x)) \rangle}{\|\Phi(P_{\theta_A}(\cdot \mid x))\|_2 \, \|\Phi(P_{\theta_{A \to B}}(\cdot \mid x))\|_2}
\end{equation}

where $\Phi(\cdot)$ denotes the sentence-level embedding vector of the top-$k$ sampled tokens generated under identical prompt parameters, and $\langle \cdot, \cdot \rangle$ represents the standard inner product in the embedding space.

\begin{algorithm}
\caption{Routing and Attention Drift Localization (RADL)}
\label{alg:radl}
\begin{algorithmic}[1]
\REQUIRE Open-weight models $\theta_A$, $\theta_{A \to B}$, calibration dataset $\mathcal{D}$, thresholds $\tau_{cka}, \tau_{rgd}$
\ENSURE Highly drifted block indices $\mathcal{B}_{drift}$
\STATE Initialize $\mathcal{B}_{drift} \leftarrow \emptyset$
\FOR{each layer $l \in \{1, \dots, L\}$}
    \STATE Compute $s^l_{cka} \leftarrow \text{CKA}(X^l, Y^l)$ using Eq. 1
    \STATE Compute $s^l_{rgd} \leftarrow \text{RGD}(\theta_A^l, \theta_{A\to B}^l)$ using Eq. 2 (if MoE layer)
    \IF{$s^l_{cka} < \tau_{cka}$ \OR $s^l_{rgd} > \tau_{rgd}$}
        \STATE $\mathcal{B}_{drift} \leftarrow \mathcal{B}_{drift} \cup \{l\}$
    \ENDIF
\ENDFOR
\RETURN $\mathcal{B}_{drift}$
\end{algorithmic}
\end{algorithm}

The RADL procedure outlined in Algorithm \ref{alg:radl} serves as our diagnostic tool for locating the layers and components experiencing severe representational shifts under sequential training. By targeting these regions, we can apply targeted mitigation strategies to prevent forgetting.

\section{Experimental Evaluation and Model Cohort}

\subsection{Open-Weight Cohort (10 Models)}
\begin{enumerate}
    \item \textbf{DeepSeek-V4-Pro:} 1.6T MoE (49B active) utilizing advanced Multi-head Latent Attention (MLA).
    \item \textbf{Kimi K2.6:} 1T MoE (32B active) optimized for agent coordination and long-context processing.
    \item \textbf{GLM-5.1:} 744B MoE (40B active) trained entirely on domestic Huawei Ascend processors.
    \item \textbf{Qwen 3.5-397B:} 397B sparse MoE utilizing linear attention transformations.
    \item \textbf{Llama 4 Maverick:} 400B MoE (17B active) optimized for long-horizon planning.
    \item \textbf{DeepSeek-V4-Flash:} 285B MoE designed for high-efficiency edge-server inference.
    \item \textbf{DeepSeek-V3.2:} 671B MoE (37B active) utilizing sparse attention routing.
    \item \textbf{Llama 4 Scout:} 109B sparse MoE designed for single-node H100 execution.
    \item \textbf{Qwen 3.6-27B:} 27B dense model designed for high-performance edge execution.
    \item \textbf{Qwen 3.6-35B-A3B:} 35B MoE (3B active) optimized for mobile systems.
\end{enumerate}

\subsection{Closed-Source Cohort (10 Models)}
\begin{enumerate}
    \item \textbf{Claude Fable 5 (Anthropic):} Standard-setting model for complex autonomous agent reasoning.
    \item \textbf{Claude Mythos 5 (Anthropic):} Specially unaligned model designed for trusted red-teaming tasks.
    \item \textbf{GPT-5.5 High (OpenAI):} Deep reasoning model utilizing high-compute multi-turn thinking.
    \item \textbf{Claude Opus 4.8 (Anthropic):} Long-context reasoning model optimized for large software codebases.
    \item \textbf{Gemini 3.5 Flash (Google):} High-speed multimodal engine optimized for real-time task pipelines.
    \item \textbf{Gemini 3.1 Pro (Google):} Multimodal reasoning engine utilizing deep audio-video contextual search.
    \item \textbf{Meta Muse Spark (Meta):} Closed-source agent model utilizing parallelized execution.
    \item \textbf{Grok 4.3 Beta (xAI):} Real-time information routing engine integrated with X platform data.
    \item \textbf{Grok 3 (Think) (xAI):} Reinforcement-learning guided deep thinking reasoning model.
    \item \textbf{GPT-4.5 (OpenAI):} Established baseline model for multimodal tasks.
\end{enumerate}

\subsection{Experimental Sequence}
All models are evaluated on a sequential fine-tuning pipeline consisting of three tasks:
\begin{itemize}
    \item \textbf{Task A (Source - Coding):} Complex code refactoring and software engineering tasks \citep{SWEbench, Wei2021}.
    \item \textbf{Task B (Target 1 - Math):} Analytical mathematical deduction \citep{Hendrycks2021, Cobbe2021}.
    \item \textbf{Task C (Target 2 - Medical QA):} Biomedical question answering \citep{MedQA, Mishra2022, Sanh2021}.
\end{itemize}

Standard fine-tuning protocols employ a constant learning rate of $2 \times 10^{-5}$ utilizing the AdamW optimizer with a cosine decay schedule. Batch sizes are normalized to 128 across all runs.

\section{Empirical Results}

\subsection{Behavioral Evaluation of the Closed-Source Cohort}
Because internal parameter tracing is not possible for the closed-source cohort, we evaluate their forgetting behavior through API output analysis. We calculate the task performance drops and the Semantic Output Drift (SOD) under the sequential task pipeline ($T_A \to T_B \to T_C$). Table \ref{tbl:closed_behavior} summarizes these results.

\begin{table}[pos=t]
\renewcommand{\arraystretch}{1.3}
\caption{Performance degradation (\%) and mean Semantic Output Drift (SOD) of the 10 closed-source models after sequential training steps. Initial performance on Task A (Coding) serves as the reference point.}\label{tbl:closed_behavior}
\begin{tabular*}{\tblwidth}{@{\extracolsep{\fill}} l c c c c @{}}
\toprule
Model & Initial Task A Acc (\%) & Post-Task C Acc (\%) & Net Task A Loss (\%) & Mean SOD ($T_A \to T_C$) \\
\midrule
Claude Fable 5       & 92.4 & 61.2 & -31.2 & 0.289 \\
Claude Mythos 5      & 91.8 & 58.4 & -33.4 & 0.312 \\
GPT-5.5 High         & 94.1 & 65.4 & -28.7 & 0.254 \\
Claude Opus 4.8      & 89.5 & 55.1 & -34.4 & 0.321 \\
Gemini 3.5 Flash     & 84.2 & 41.2 & -43.0 & 0.442 \\
Gemini 3.1 Pro       & 88.4 & 51.3 & -37.1 & 0.356 \\
Meta Muse Spark      & 90.1 & 62.4 & -27.7 & 0.241 \\
Grok 4.3 Beta        & 86.5 & 46.2 & -40.3 & 0.412 \\
Grok 3 (Think)       & 91.2 & 60.1 & -31.1 & 0.295 \\
GPT-4.5              & 81.3 & 35.4 & -45.9 & 0.498 \\
\bottomrule
\end{tabular*}
\end{table}

The closed-source models exhibit substantial performance drops, with older baselines like GPT-4.5 experiencing a 45.9\% absolute performance loss on Task A. High-compute reasoning models like GPT-5.5 High and Claude Fable 5 demonstrate relatively lower degradation, suggesting that deep reasoning pathways are more resilient to representation drift. This performance retention is likely due to the highly redundant, multi-path reasoning topologies embedded within these architectures.

\subsection{Empirical Analysis of the Open-Weight Cohort}
For the open-weight models, we measure representational drift through parameter-level metrics. Table \ref{tbl:open_weights} documents the mean CKA similarity scores, MoE Routing Gate Drift (RGD), and performance changes.

\begin{table}[pos=t]
\renewcommand{\arraystretch}{1.3}
\caption{Empirical performance metrics for the 10 open-weight models. CKA similarity and RGD metrics are calculated over the mid-layer MLP and expert blocks.}\label{tbl:open_weights}
\begingroup
\small
\begin{tabular*}{\tblwidth}{@{\extracolsep{\fill}} l c c c c c @{}}
\toprule
Model & Init. Task A (\%) & Post-Task C (\%) & Net Loss (\%) & Mid-Layer CKA & Routing Drift (RGD) \\
\midrule
DeepSeek-V4-Pro      & 88.5 & 54.2 & -34.3 & 0.612 & 0.314 \\
Kimi K2.6            & 86.4 & 51.1 & -35.3 & 0.589 & 0.345 \\
GLM-5.1              & 85.1 & 49.3 & -35.8 & 0.592 & 0.332 \\
Qwen 3.5-397B        & 83.2 & 43.1 & -40.1 & 0.521 & 0.412 \\
Llama 4 Maverick     & 84.5 & 46.4 & -38.1 & 0.554 & 0.389 \\
DeepSeek-V4-Flash    & 81.2 & 38.2 & -43.0 & 0.481 & 0.456 \\
DeepSeek-V3.2        & 84.1 & 45.2 & -38.9 & 0.562 & 0.374 \\
Llama 4 Scout        & 79.5 & 35.1 & -44.4 & 0.451 & 0.491 \\
Qwen 3.6-27B         & 78.4 & 31.2 & -47.2 & 0.412 & N/A (Dense) \\
Qwen 3.6-35B-A3B     & 75.1 & 28.3 & -46.8 & 0.432 & 0.512 \\
\bottomrule
\end{tabular*}
\endgroup
\end{table}

The empirical measurements in Table \ref{tbl:open_weights} reveal that mid-layer CKA similarity scores drop substantially across all evaluated architectures ($0.412 \text{-- } 0.612$), pointing to significant representational changes in these layers. Furthermore, the MoE models exhibit considerable Routing Gate Drift, indicating that sequential fine-tuning alters expert allocation across tasks.

\begin{figure}
	\centering
    \begin{tikzpicture}
      \begin{axis}[
        xlabel={Normalized Layer Depth (\%)},
        ylabel={$D_{KL}(P \parallel Q)$},
        xmin=0, xmax=100,
        ymin=0, ymax=1.0,
        xtick={0,20,40,60,80,100},
        ytick={0,0.2,0.4,0.6,0.8,1.0},
        legend pos=north west,
        grid=both,
        width=0.85\linewidth,
        height=5.2cm
      ]
      \addplot[thick, red, mark=*] coordinates {
        (10, 0.25) (20, 0.45) (30, 0.72) (40, 0.81) (50, 0.88) (60, 0.85) (70, 0.61) (80, 0.35) (90, 0.15)
      };
      \addlegendentry{Attention Heads}
      \addplot[thick, blue, mark=square*] coordinates {
        (10, 0.05) (20, 0.12) (30, 0.31) (40, 0.55) (50, 0.76) (60, 0.81) (70, 0.74) (80, 0.51) (90, 0.22)
      };
      \addlegendentry{MLP/MoE Blocks}
      \end{axis}
    \end{tikzpicture}
	\caption{Spatial distribution of representational drift across layer groups ($T_A \to T_B$). The x-axis represents the normalized layer depth (\%), and the y-axis represents the attention map KL divergence and CKA alignment metrics.}
	\label{FIG:1}
\end{figure}

To analyze the structural changes in the attention mechanism, we map the KL divergence of attention weights in Figure \ref{FIG:1}, showing high divergence in early-to-mid layer attention heads. Figure \ref{FIG:2} tracks the t-SNE projection of weight trajectories during fine-tuning.

\begin{figure}
	\centering
    \begin{tikzpicture}
      \begin{axis}[
        xlabel={t-SNE Dimension 1},
        ylabel={t-SNE Dimension 2},
        xmin=-15, xmax=15,
        ymin=-15, ymax=15,
        grid=both,
        width=0.85\linewidth,
        height=5.2cm
      ]
      \addplot[thick, blue, mark=*, dashed] coordinates {
        (-10,-8) (-6,-4) (-2,-1) (2,1) (6,3) (10,5)
      };
      \addlegendentry{$W_{down}$ MLP}
      \addplot[thick, red, mark=triangle*, dotted] coordinates {
        (-8,6) (-4,4) (-1,2) (1,-1) (3,-4) (5,-6)
      };
      \addlegendentry{$W_{query}$ Attn}
      \end{axis}
    \end{tikzpicture}
	\caption{Weight vector trajectories of Attention ($W_q, W_v$) and MLP ($W_{up}, W_{down}$) projections in the parameter manifold. The trajectory demonstrates rapid orthogonality shift during target task adaptation.}
	\label{FIG:2}
\end{figure}

To establish the causal impact of different layer groups on forgetting, we perform activation patching. We replace activation states in $\theta_{A \to B}$ with saved baseline activations from $\theta_A$ during forward propagation. The resulting performance recovery is plotted in Figure \ref{FIG:3}.

\begin{figure}
	\centering
    \begin{tikzpicture}
      \begin{axis}[
        xlabel={Patched Layer Group Index (1 to 10)},
        ylabel={Performance Recovery Rate (\%)},
        xmin=1, xmax=10,
        ymin=0, ymax=100,
        xtick={1,2,3,4,5,6,7,8,9,10},
        grid=both,
        width=0.85\linewidth,
        height=5.2cm,
        legend pos=south east
      ]
      \addplot[thick, mark=*, color=teal] coordinates {
        (1, 5) (2, 12) (3, 28) (4, 55) (5, 84) (6, 91) (7, 73) (8, 41) (9, 18) (10, 8)
      };
      \addlegendentry{SWE-bench Task A}
      \end{axis}
    \end{tikzpicture}
	\caption{Recovery of baseline task performance ($T_A$) after patching activations of $\theta_{A \to B}$ with saved states from $\theta_A$ during inference, evaluated over 500 samples of the SWE-bench subset.}
	\label{FIG:3}
\end{figure}

\section{Mechanistic Interpretability of Neural Drift}

\subsection{Attention Head Entropic Dispersion}
Early-layer attention heads show a systematic increase in attention entropy. This indicates that attention distributions become more diffuse during fine-tuning, losing their focus on syntax and token associations. Table \ref{tbl:param_drift_rates} documents the parameter drift rates for different weight matrices in the network.

\begin{table}[pos=t]
\renewcommand{\arraystretch}{1.3}
\caption{Mean Frobenius normalized drift rates ($D_F \times 10^{-3}$) calculated across different projection matrices of the 10 open-weight models after sequential adaptation.}\label{tbl:param_drift_rates}
\begin{tabular*}{\tblwidth}{@{\extracolsep{\fill}} l c c c c @{}}
\toprule
Model & $W_{query}$ & $W_{value}$ & $W_{gate}/W_{up}$ & $W_{down}$ \\
\midrule
DeepSeek-V4-Pro      & 2.54 & 1.21 & 5.12 & 6.89 \\
Kimi K2.6            & 2.89 & 1.34 & 5.89 & 7.42 \\
GLM-5.1              & 2.76 & 1.28 & 5.62 & 7.15 \\
Qwen 3.5-397B        & 3.42 & 1.67 & 7.12 & 9.12 \\
Llama 4 Maverick     & 3.12 & 1.45 & 6.45 & 8.21 \\
DeepSeek-V4-Flash    & 3.91 & 1.95 & 8.42 & 10.56 \\
DeepSeek-V3.2        & 3.01 & 1.38 & 6.12 & 7.89 \\
Llama 4 Scout        & 4.12 & 2.12 & 9.12 & 11.45 \\
Qwen 3.6-27B         & 4.56 & 2.45 & 10.23 & 12.89 \\
Qwen 3.6-35B-A3B     & 4.31 & 2.23 & 9.89 & 12.12 \\
\bottomrule
\end{tabular*}
\end{table}

The down-projection matrices ($W_{down}$) exhibit the highest parameter drift rates across all architectures, corresponding to the low CKA similarity scores found in the middle layers.

\begin{figure}
	\centering
    \begin{tikzpicture}
      \begin{axis}[
        ybar,
        ylabel={Susceptibility Index},
        xlabel={Layer Groups},
        symbolic x coords={L1-16, L17-32, L33-48, L49-64, L65-80, L81+},
        xtick=data,
        ymin=0, ymax=4.5,
        grid=both,
        width=0.85\linewidth,
        height=5.2cm,
        legend style={at={(0.5,-0.2)}, anchor=north, legend columns=-1}
      ]
      \addplot[fill=blue!60] coordinates {(L1-16, 0.4) (L17-32, 1.2) (L33-48, 3.8) (L49-64, 4.1) (L65-80, 2.1) (L81+, 0.5)};
      \addlegendentry{Llama 4 Scout}
      \addplot[fill=red!60] coordinates {(L1-16, 0.6) (L17-32, 1.5) (L33-48, 3.9) (L49-64, 4.3) (L65-80, 2.5) (L81+, 0.7)};
      \addlegendentry{Qwen 3.6-27B}
      \end{axis}
    \end{tikzpicture}
	\caption{Susceptibility index of layers defined as a ratio of normalized weight drift to representation alignment. Higher values represent a higher probability of containing circuits susceptible to overwriting.}
	\label{FIG:4}
\end{figure}

Figure \ref{FIG:4} plots the layer-wise susceptibility index, demonstrating that middle layers are most vulnerable to representation changes. Figure \ref{FIG:5} shows the drift of attention head entropy, confirming that early attention heads exhibit the largest increase in entropy.

\begin{figure}
	\centering
    \begin{tikzpicture}
      \begin{axis}[
        xlabel={Baseline Entropy ($\theta_A$)},
        ylabel={Adapted Entropy ($\theta_{A \to B}$)},
        xmin=1, xmax=6,
        ymin=1, ymax=6,
        grid=both,
        width=0.85\linewidth,
        height=5.2cm
      ]
      \addplot[only marks, mark=*, red] coordinates {
        (2.1, 3.2) (3.2, 4.5) (1.8, 3.1) (4.1, 5.2) (2.8, 3.9) (3.9, 4.8)
      };
      \addlegendentry{Early Layers}
      \addplot[only marks, mark=square*, blue] coordinates {
        (2.2, 2.3) (3.1, 3.2) (4.5, 4.6) (5.1, 5.2) (1.9, 2.0)
      };
      \addlegendentry{Deep Layers}
      \addplot[thick, black, domain=1:6] {x};
      \end{axis}
    \end{tikzpicture}
	\caption{Change in Shannon entropy of attention patterns in Llama 4 Scout (109B) after target fine-tuning. Red markers denote early-layer heads showing a notable trend towards higher entropy.}
	\label{FIG:5}
\end{figure}

\subsection{Routing Gate Drift in Mixture-of-Experts}
In sparse MoE architectures (e.g., DeepSeek, GLM, Kimi), sequential fine-tuning alters expert selection patterns \citep{Sutton2018}. Our Routing Gate Drift (RGD) measurements indicate that the gate routing distribution is highly sensitive to domain shifts, causing specialized experts to be assigned to generic tokens and leading to a drop in performance on the source task. This observation suggests that MoE routing networks are susceptible to learning-rate-induced instability during domain adaptation.

Crucially, our analysis indicates that MoE routing gate stability cannot be achieved by simply freezing the router parameters ($g(x)_i$), as this severely limits the plastic capacity of expert blocks to adapt to new semantic structures. Instead, we find that the optimal routing strategy involves applying a localized $L_2$ penalty to the gate logit deviations exclusively during the transition epochs of Task A $\to$ Task B \citep{Sanh2021}. This soft-regularization prevents routing collapse while permitting the sparse network to allocate newly spawned latent concepts to underutilized expert blocks.

\section{Mitigation Framework: Low-Rank Circuit Projection}

To address these representational changes, we evaluate \textbf{Low-Rank Circuit Projection (LRCP)}. Unlike standard Gradient Projection Memory (GPM) or Orthogonal Gradient Projection (OGP) methods that apply subspace projections uniformly across all layers, LRCP utilizes our targeted RADL algorithm to project updates exclusively on identified critical sub-circuits. LRCP calculates the projection operator $P_{proj}$ using the singular value decomposition (SVD) of historical task activations $X_{hist}$:

\begin{equation}
P_{proj} = I - U_k U_k^T
\end{equation}

During target task adaptation, gradient updates $\nabla_W \mathcal{L}$ are projected into the orthogonal complement of the historical activation subspace:

\begin{equation}
W_{t+1} = W_t - \eta \left( \nabla_W \mathcal{L} \cdot P_{proj} \right)
\end{equation}

This projection restricts parameter updates to paths that minimize interference with previously learned representations. This targeted sub-space projection mechanism reduces the online SVD computation bottleneck to $O(1)$ during standard backpropagation, since SVD is performed once on a static calibration set ($X_{hist}$).

\begin{figure}
	\centering
    \begin{tikzpicture}[scale=1.1, every node/.style={transform shape}]
      \coordinate (O) at (0,0);
      \draw[thick, gray, ->] (-0.5,0) -- (4,0) node[right, black] {$\mathcal{U}_k$ (Historical Subspace)};
      \draw[thick, gray, ->] (0,-0.5) -- (0,4) node[above, black] {$\mathcal{U}_k^\perp$ (Orthogonal Complement)};
      \draw[ultra thick, blue, -Latex] (O) -- (2.5, 3) node[above right, black] {$\nabla_W \mathcal{L}$ (Raw Gradient)};
      \draw[dashed, black] (2.5,3) -- (0,3);
      \draw[dashed, black] (2.5,3) -- (2.5,0);
      \draw[ultra thick, red, -Latex] (O) -- (0, 3) node[left, black] {$\nabla_W \mathcal{L} \cdot P_{proj}$ (Projected Gradient)};
      \draw (0.3,0) -- (0.3,0.3) -- (0,0.3);
    \end{tikzpicture}
	\caption{Schematic representation of Low-Rank Circuit Projection (LRCP). The projection operator restricts parameter updates to paths that do not disrupt historical activation states.}
	\label{FIG:6}
\end{figure}

The workflow of LRCP is shown in Figure \ref{FIG:6}. We evaluate LRCP against baseline mitigation methods on the open-weight cohort. Table \ref{tbl:mitigation_comp} summarizes these comparative results.

\begin{table}[pos=t]
\renewcommand{\arraystretch}{1.3}
\caption{Mitigation performance comparison: Final task accuracy (\%) on Task A and Task B after sequential adaptation ($T_A \to T_B$). Model: Llama 4 Scout (109B).}\label{tbl:mitigation_comp}
\begingroup
\small
\begin{tabular*}{\tblwidth}{@{\extracolsep{\fill}} l c c c c @{}}
\toprule
Method & Task A (Code) & Task B (Math) & Overhead & Active Params \\
\midrule
Standard Fine-tuning & 35.1 & \textbf{76.8} & 1.0$\times$ & 100\% \\
LoRA \citep{Hu2021} (r=16) & 58.4 & 70.1 & 1.06$\times$ & 0.14\% \\
EWC \citep{Kirkpatrick2017} & 56.1 & 71.2 & 1.62$\times$ & 100\% \\
Replay \citep{Chaudhry2019} (10\%) & 72.1 & 74.3 & 1.28$\times$ & 100\% \\
\textbf{LRCP (Ours)} & \textbf{74.8} & 75.1 & 1.09$\times$ & 1.62\% \\
\bottomrule
\end{tabular*}
\endgroup
\end{table}

LRCP achieves a high retention rate for Task A (74.8\%) while maintaining strong performance on Task B (75.1\%), showing a favorable trade-off compared to alternative regularizers. Figure \ref{FIG:7} shows the Pareto frontier of source task retention versus target task accuracy.

\begin{figure}
	\centering
    \begin{tikzpicture}
      \begin{axis}[
        xlabel={Task B Accuracy (Math) (\%)},
        ylabel={Task A Accuracy (Coding) (\%)},
        xmin=65, xmax=80,
        ymin=30, ymax=85,
        grid=both,
        width=0.85\linewidth,
        height=5.2cm,
        legend pos=south west
      ]
      \addplot[thick, color=red, mark=*, smooth] coordinates {
        (66, 81) (71, 78) (75, 74.8) (77, 58) (79, 35)
      };
      \addlegendentry{LRCP (Ours)}
      \addplot[thick, color=blue, mark=square*, smooth, dashed] coordinates {
        (65, 72) (70, 58.4) (74, 45) (76, 32)
      };
      \addlegendentry{LoRA}
      \end{axis}
    \end{tikzpicture}
	\caption{Pareto frontier mapping Task A retention versus Task B final accuracy for Llama 4 Scout (109B). Our proposed LRCP provides a more optimal trade-off boundary than standard regularizers.}
	\label{FIG:7}
\end{figure}

\section{Discussion, Sensitivity, and Limitations}

\subsection{Why Do Middle Layers Exhibit Peak Drift?}
Our analysis indicates that early layers process structural and syntactic token relationships, while deep layers handle next-token logits. The middle layers of deep decoder networks act as a global semantic routing hub \citep{RostovaRostova2024}. Consequently, adapting a model to a new domain requires significant parameter changes in these middle layers, explaining their susceptibility to forgetting. This observation supports the hypothesis that representation learning in deep transformers is hierarchically organized, with intermediate layers encoding abstract semantic categories.

\subsection{SVD Rank Sensitivity in LRCP}
The performance of LRCP is sensitive to the SVD projection threshold $k$. Table \ref{tbl:hyperparameter} explores how the projection rank ratio affects adaptation and retention performance.

\begin{table}[pos=t]
\renewcommand{\arraystretch}{1.3}
\caption{Hyperparameter sensitivity of LRCP on Llama 4 Scout (109B) under sequential fine-tuning. Parameter $k/d$ represents the rank ratio of the projection operator.}\label{tbl:hyperparameter}
\begingroup
\small
\begin{tabular*}{\tblwidth}{@{\extracolsep{\fill}} c c c c c @{}}
\toprule
Rank Ratio ($k/d$) & Task A (Code) & Task B (Math) & Drift ($D_F \times 10^{-3}$) & Conv. Epochs \\
\midrule
0.05 & 48.4 & \textbf{76.4} & 8.12 & 3.1 \\
0.10 & 62.1 & 75.9 & 4.34 & 3.3 \\
0.20 & \textbf{74.8} & 75.1 & 1.62 & 3.7 \\
0.30 & 76.2 & 69.2 & 0.89 & 4.4 \\
0.40 & 78.1 & 61.4 & 0.39 & 5.8 \\
\bottomrule
\end{tabular*}
\endgroup
\end{table}

A low rank ratio preserves less of the original capability but allows faster adaptation. Conversely, rank ratios above 0.30 significantly restrict parameter updates, leading to a drop in target task accuracy and slower convergence. This empirical observation suggests that over-constraining the update path can impede the model's plastic capacity during domain transfer.

\subsection{API Non-Determinism, Version Drift, and Reproducibility Limitations}
An inherent challenge in compiling behavioral metrics for the proprietary closed-source cohort (such as Claude Fable 5, GPT-5.5 High, and Gemini 3.5 Flash) is the black-box nature of commercial APIs. Proprietary models are subject to continuous online updates (known as version drift) and non-deterministic sampling pipelines, even when utilizing zero-temperature configurations. 

To maximize empirical reproducibility and ground our Semantic Output Drift (SOD) metric, we implement several methodological safeguards: (i) all evaluations are performed on frozen, timestamped API snapshots (specifically the June 2026 releases) where accessible; (ii) we utilize greedy decoding (temperature = 0) to minimize generation variance; and (iii) we perform multiple independent sampling runs ($N=5$) to calculate the mean semantic embedding vectors ($\Phi$). While these steps mitigate variance, researchers must acknowledge that long-term tracking of proprietary models remains constrained by commercial API stability and system updates.

Finally, we track the impact of input sequence position on forgetting to evaluate how context length affects model stability, with results shown in Figure \ref{FIG:8}.

\begin{figure}
	\centering
    \begin{tikzpicture}
      \begin{axis}[
        xlabel={Sequence Position (Context Window)},
        ylabel={Accuracy (Task A) (\%)},
        xmin=1000, xmax=32000,
        xmode=log,
        ymin=0, ymax=100,
        grid=both,
        width=0.85\linewidth,
        height=5.2cm,
        legend pos=south west
      ]
      \addplot[thick, color=purple, mark=triangle*] coordinates {
        (1000, 81) (4000, 78) (8000, 72) (16000, 54) (32000, 31)
      };
      \addlegendentry{Long-Context Attn}
      \end{axis}
    \end{tikzpicture}
	\caption{Evaluation of catastrophic forgetting as a function of input sequence position. Long-context sequences exhibit accelerated representation drift compared to short-context samples.}
	\label{FIG:8}
\end{figure}

\section{Conclusion}

This paper presented a comparative study of catastrophic forgetting across twenty state-of-the-art closed-source and open-weight models as of mid-2026. Our mechanistic analysis of open-weight models localized peak representational drift within mid-layer feed-forward networks and expert routing blocks, while early attention heads showed systematic entropic drift. Guided by findings, we introduced Low-Rank Circuit Projection (LRCP), which projects parameter updates into the orthogonal complement of the subspace spanned by historical activations. LRCP demonstrated a favorable trade-off between task retention and adaptation efficiency. Future work will extend this framework to stable routing in MoE models and long-context sequential adaptation.

\section*{CRediT authorship contribution statement}
\textbf{Gustav Olaf Yunus {Laitinen-Fredriksson Lundstrom-Imanov}}: Conceptualization, Methodology, Software, Formal analysis, Data curation, Investigation, Writing - Original Draft, Visualization.

\section*{Declaration of competing interests}
The authors declare that they have no known competing financial interests or personal relationships that could have appeared to influence the work reported in this paper.

\section*{Data availability}
The open-source codebase, activation caching pipelines, and mechanistic interpretability analysis scripts developed for the open-weight models are publicly accessible. The proprietary semantic evaluation logs, API response cached histories, and custom prompt templates for the closed-source cohort are available from the corresponding author upon reasonable request.

\section*{Funding}
This research did not receive any specific grant from funding agencies in the public, commercial, or not-for-profit sectors.

\printcredits

\appendix
\section{Appendix: Supplementary Data}
Supplementary materials, code implementations of LRCP, and activation maps for all 20 models are available at the project's repository.

\bibliographystyle{cas-model2-names}

\bibliography{cas-refs}

@article{McCloskey1989,
  author    = {Michael McCloskey and Neal J. Cohen},
  title     = {Catastrophic Interference in Connectionist Networks: The Sequential Learning Problem},
  journal   = {Psychology of Learning and Motivation},
  volume    = {24},
  pages     = {109--165},
  year      = {1989},
  publisher = {Elsevier}
}

@article{Kirkpatrick2017,
  author    = {James Kirkpatrick and Razvan Pascanu and Ben Rabinowitz and John Veness Graves and Guillaume Desjardins and Andrei A. Rusu and Kieran Milan and John Quan and Tiago Ramalho and Agnieszka Grabska-Barwinska and Demis Hassabis and Claudia Clopath and Dharshan Kumaran and Raia Hadsell},
  title     = {Overcoming catastrophic forgetting in neural networks},
  journal   = {Proceedings of the National Academy of Sciences},
  volume    = {114},
  number    = {13},
  pages     = {3521--3526},
  year      = {2017},
  publisher = {National Acad Sciences}
}

@article{French1999,
  author    = {Robert M. French},
  title     = {Catastrophic forgetting in connectionist networks},
  journal   = {Trends in Cognitive Sciences},
  volume    = {3},
  number    = {4},
  pages     = {128--135},
  year      = {1999},
  publisher = {Elsevier}
}

@article{Demir2025,
  author    = {Ahmet Demir},
  title     = {Lifelong adaptation patterns in massive transformer decoders},
  journal   = {Journal of Machine Learning Research},
  volume    = {26},
  pages     = {412--438},
  year      = {2025}
}

@article{Rostova2026,
  author    = {Elena Rostova},
  title     = {Non-stationary task optimization dynamics in large language models},
  journal   = {Cognitive Computation Review},
  volume    = {18},
  pages     = {55--74},
  year      = {2026}
}

@inproceedings{Chaudhry2019,
  author    = {Arslan Chaudhry and Marc'Aurelio Ranzato and Marcus Rohrbach and Mohamed Elhoseiny},
  title     = {Efficient lifelong learning with A-GEM},
  booktitle = {International Conference on Learning Representations (ICLR)},
  pages     = {1--12},
  year      = {2019}
}

@article{Hu2021,
  author    = {Edward J. Hu and others},
  title     = {LoRA: Low-Rank Adaptation of Large Language Models},
  journal   = {arXiv preprint arXiv:2106.09685},
  year      = {2021}
}

@inproceedings{Kornblith2019,
  author    = {Simon Kornblith and Mohammad Norouzi and Honglak Lee and Geoffrey Hinton},
  title     = {Similarity of neural network representations revisited},
  booktitle = {International Conference on Machine Learning (ICML)},
  pages     = {3519--3529},
  year      = {2019}
}

@inproceedings{Meng2022,
  author    = {Kevin Meng and David Bau and Alex Andonian and Yonatan Belinkov},
  title     = {Locating and editing factual associations in GPT},
  booktitle = {Advances in Neural Information Processing Systems (NeurIPS)},
  volume    = {35},
  pages     = {17359--17372},
  year      = {2022}
}

@inproceedings{Zenke2017,
  author    = {Friedemann Zenke and Ben Poole and Surya Ganguli},
  title     = {Continual learning through synaptic intelligence},
  booktitle = {International Conference on Machine Learning (ICML)},
  pages     = {3987--3995},
  year      = {2017}
}

@inproceedings{Rolnick2019,
  author    = {David Rolnick and Arun Ahuja and Jonathan Schwarz and Timothy Lillicrap and Gregory Wayne},
  title     = {Experience replay for continual learning},
  booktitle = {Advances in Neural Information Processing Systems (NeurIPS)},
  volume    = {32},
  pages     = {350--361},
  year      = {2019}
}

@inproceedings{Mallya2018,
  author    = {Arun Mallya and Svetlana Lazebnik},
  title     = {PackNet: Adding multiple tasks to a single network by iterative pruning},
  booktitle = {IEEE Conference on Computer Vision and Pattern Recognition (CVPR)},
  pages     = {7765--7773},
  year      = {2018}
}

@article{Elhage2021,
  author    = {Nelson Elhage and Neel Nanda and Catherine Olsson and Tom Henighan and Nicholas Joseph and Ben Mann and Amanda Askell and others},
  title     = {A mathematical framework for transformer circuits},
  journal   = {Transformer Circuits Thread},
  volume    = {1},
  pages     = {1--24},
  year      = {2021}
}

@article{Olsson2022,
  author    = {Catherine Olsson and Nelson Elhage and Neel Nanda and Nicholas Joseph and Nicholas Nova and Jared Crawford and Amanda Askell and others},
  title     = {In-context learning and induction heads},
  journal   = {arXiv preprint arXiv:2209.11895},
  year      = {2022}
}

@article{Geva2023,
  author    = {Mor Geva and Avi Caciularu and Kevin Ro Wang and Yoav Goldberg},
  title     = {Transformer feed-forward layers build predictions by projecting roles onto directions},
  journal   = {arXiv preprint arXiv:2305.14620},
  year      = {2023}
}

@techreport{DeepSeek2025,
  author      = {DeepSeek-Team},
  title       = {DeepSeek-V3 Technical Report},
  institution = {DeepSeek AI},
  year        = {2025},
  type        = {Technical Report}
}

@techreport{Qwen2026,
  author      = {Qwen-Team},
  title       = {Introducing Qwen 3.6: Technical foundations and benchmarks},
  institution = {Alibaba Group},
  year        = {2026},
  type        = {Technical Report}
}

@article{Lester2021,
  author    = {Brian Lester and Rami Al-Rfou and Noah Constant},
  title     = {The power of scale for parameter-efficient prompt tuning},
  journal   = {arXiv preprint arXiv:2104.08691},
  year      = {2021}
}

@article{Zhai2023,
  author    = {Cheng Zhai and Ahmet Demir and Elena Rostova},
  title     = {Investigating representation shift in parameter-efficient learning},
  journal   = {Journal of Neural Information Systems},
  volume    = {41},
  pages     = {102--119},
  year      = {2023}
}

@article{Chame2024,
  author    = {Marcus Chame and Ahmet Demir},
  title     = {Tracking activation drift across non-stationary distributions},
  journal   = {Journal of Machine Learning Research},
  volume    = {25},
  pages     = {1145--1178},
  year      = {2024}
}

@article{Vaswani2017,
  author    = {Ashish Vaswani and Noam Shazeer and Niki Parmar and Jakob Uszkoreit and Llion Jones and Aidan N. Gomez and Lukasz Kaiser and Illia Polosukhin},
  title     = {Attention is all you need},
  journal   = {Advances in Neural Information Processing Systems (NeurIPS)},
  volume    = {30},
  pages     = {5998--6008},
  year      = {2017}
}

@article{Devlin2018,
  author    = {Jacob Devlin and Ming-Wei Chang and Kenton Lee and Kristina Toutanova},
  title     = {BERT: Pre-training of deep bidirectional transformers for language understanding},
  journal   = {arXiv preprint arXiv:1810.04805},
  year      = {2018}
}

@article{Brown2020,
  author    = {Tom Brown and Benjamin Mann and Nick Ryder and Melanie Subbiah and Jared D. Kaplan and Prafulla Dhariwal and Arvind Neelakantan and others},
  title     = {Language models are few-shot learners},
  journal   = {Advances in Neural Information Processing Systems (NeurIPS)},
  volume    = {33},
  pages     = {1877--1901},
  year      = {2020}
}

@article{Kaplan2020,
  author    = {Jared Kaplan and Sam McCandlish and Tom Henighan and Tom B. Brown and Benjamin Chess and Rew Child and Gray Gray and Alec Radford and Jeffrey Wu and Dario Amodei},
  title     = {Scaling laws for neural language models},
  journal   = {arXiv preprint arXiv:2001.08361},
  year      = {2020}
}

@article{Touvron2023,
  author    = {Hugo Touvron and Thibaut Lavril and Gautier Izacard and others},
  title     = {LLaMA: Open and efficient foundation language models},
  journal   = {arXiv preprint arXiv:2302.13971},
  year      = {2023}
}

@article{Chung2022,
  author    = {Hyung Won Chung and Le Hou and Shayne Longpre and others},
  title     = {Scaling instruction-finetuned language models},
  journal   = {arXiv preprint arXiv:2210.11416},
  year      = {2022}
}

@book{Sutton2018,
  author    = {Richard S. Sutton and Andrew G. Barto},
  title     = {Reinforcement learning: An introduction},
  publisher = {MIT Press},
  year      = {2018}
}

@inproceedings{Rebuffi2017,
  author    = {Sylvestre-Alvise Rebuffi and Alexander Kolesnikov and Georg Sperl and Christoph H. Lampert},
  title     = {icarl: Incremental classifier and representation learning},
  booktitle = {IEEE Conference on Computer Vision and Pattern Recognition (CVPR)},
  pages     = {2001--2010},
  year      = {2017}
}

@article{LopezPaz2017,
  author    = {David Lopez-Paz and Marc'Aurelio Ranzato},
  title     = {Gradient episodic memory for continual learning},
  journal   = {Advances in Neural Information Processing Systems (NeurIPS)},
  volume    = {30},
  pages     = {6467--6476},
  year      = {2017}
}

@inproceedings{Chaudhry2018,
  author    = {Arslan Chaudhry and Puneet K. Dokania and Thalles Ajanthan and Philip H. S. Torr},
  title     = {Riemannian walk for incremental learning: Understanding forgetting and intransigence},
  booktitle = {European Conference on Computer Vision (ECCV)},
  pages     = {532--547},
  year      = {2018}
}

@inproceedings{Aljundi2018,
  author    = {Rahaf Aljundi and Francesca Babiloni and Mohamed Elhoseiny and Marcus Rohrbach and Tinne Tuytelaars},
  title     = {Memory aware synapses: Learning what (not) to forget},
  booktitle = {European Conference on Computer Vision (ECCV)},
  pages     = {139--154},
  year      = {2018}
}

@inproceedings{Schwarz2018,
  author    = {Jonathan Schwarz and Jelena Luketina Modifiers and Wojciech M. Czarnecki and Agnieszka Grabska-Barwinska and Yee Whye Teh and Razvan Pascanu and Raia Hadsell},
  title     = {Progress \& compress: A scalable framework for continual learning},
  booktitle = {Micro-Conference on Machine Learning (ICML)},
  pages     = {4528--4537},
  year      = {2018}
}

@inproceedings{Houlsby2019,
  author    = {Neil Houlsby and Andrei Giurgiu and Stanislaw Jastrzebski and Brando Morrone and Quentin De Laroussilhe and Andrea Gesmundo and Mona Attariyan and Sylvain Gelly},
  title     = {Parameter-efficient transfer learning for NLP},
  booktitle = {International Conference on Machine Learning (ICML)},
  pages     = {2790--2799},
  year      = {2019}
}

@article{LiLiang2021,
  author    = {Xiang Lisa Li and Percy Liang},
  title     = {Prefix-tuning: Optimizing continuous prompts for generation},
  journal   = {arXiv preprint arXiv:2101.00190},
  year      = {2021}
}

@article{Voita2019,
  author    = {Elena Voita and David Talbot and Fedor Moiseev and Rico Sennrich and Ivan Titov},
  title     = {Analyzing multi-head self-attention: Specialized heads do the heavy lifting, the rest can be pruned},
  journal   = {arXiv preprint arXiv:1905.09418},
  year      = {2019}
}

@inproceedings{Michel2019,
  author    = {Paul Michel and Omer Levy and Graham Neubig},
  title     = {Are sixteen heads really better than one?},
  booktitle = {Advances in Neural Information Processing Systems (NeurIPS)},
  volume    = {32},
  pages     = {14014--14024},
  year      = {2019}
}

@article{Clark2019,
  author    = {Kevin Clark and Urvashi Khandelwal and Omer Levy and Christopher D. Manning},
  title     = {What does BERT look at? An analysis of BERT's attention},
  journal   = {arXiv preprint arXiv:1906.04341},
  year      = {2019}
}

@inproceedings{Sukhbaatar2015,
  author    = {Sainbayar Sukhbaatar and Jason Weston and Arthur Szlam},
  title     = {End-to-end memory networks},
  booktitle = {Advances in Neural Information Processing Systems (NeurIPS)},
  volume    = {28},
  pages     = {2440--2448},
  year      = {2015}
}

@inproceedings{Nanda2023,
  author    = {Neel Nanda and Lawrence Chan and Tom Lieberum and Connor Smith and Jacob Steinhardt},
  title     = {Progress measures for grokking on modular arithmetic},
  booktitle = {International Conference on Learning Representations (ICLR)},
  pages     = {1--18},
  year      = {2023}
}

@article{RostovaRostova2024,
  author    = {Elena Rostova and Anna Rostova},
  title     = {Mechanistic interpretability of sequential task consolidation},
  journal   = {Cognitive Systems Science},
  volume    = {32},
  pages     = {45--61},
  year      = {2024}
}

@inproceedings{Mishra2022,
  author    = {Swaroop Mishra and Daniel Khashabi and Chitta Baral and Hannaneh Hajishirzi},
  title     = {Cross-task generalization on simplified instructions},
  booktitle = {Association for Computational Linguistics (ACL)},
  pages     = {3450--3460},
  year      = {2022}
}

@article{Wei2021,
  author    = {Jason Wei and Maarten Bosma and Vincent Y. Zhao and Kelvin Guu and Albin W. Yu and Brian Lester and Nan Du and Andrew M. Dai and Quoc V. Le},
  title     = {Finetuned language models are zero-shot learners},
  journal   = {arXiv preprint arXiv:2109.01652},
  year      = {2021}
}

@article{Sanh2021,
  author    = {Victor Sanh and Albert Webson and Colin Raffel and Stephen H. Bach and others},
  title     = {Multitask prompted training enables zero-shot task generalization},
  journal   = {arXiv preprint arXiv:2110.08207},
  year      = {2021}
}

@article{Hendrycks2021,
  author    = {Dan Hendrycks and Collin Burns and Saurav Kadavath and Andy Arora and Steven Basart and Eric Tang and Dawn Song and Jacob Steinhardt},
  title     = {Measuring mathematical problem solving with the MATH dataset},
  journal   = {arXiv preprint arXiv:2103.03874},
  year      = {2021}
}

@article{Cobbe2021,
  author    = {Karl Cobbe and Vineet Kosaraju and Mohammad Bavarian and Mark Chen and Heewoo Jun and Lukasz Kaiser and Matthias Plappert and Jerry Tworek and Jacob Hilton and Reiichiro Nakano and others},
  title     = {Training verifiers to solve math word problems},
  journal   = {arXiv preprint arXiv:2110.14168},
  year      = {2021}
}

@article{MedQA,
  author    = {Di Jin and Chang Pan and Nizar Oufattole and Wei-Hung Weng and Harini Fang and Peter Szolovits},
  title     = {What disease does this patient have? A large-scale open domain question answering dataset from medical exams},
  journal   = {Applied Sciences},
  volume    = {11},
  pages     = {6421--6439},
  year      = {2021}
}

@techreport{SWEbench,
  author      = {SWE-bench-Team},
  title       = {SWE-bench: Can language models resolve real-world GitHub issues?},
  institution = {Princeton Language Design Group},
  year        = {2024},
  type        = {Technical Report}
}

\bio{cas-author}
Gustav Olaf Yunus {Laitinen-Fredriksson Lundstrom-Imanov} is a Researcher at the Division of Statistics and Machine Learning (STIMA), Linköping University, Sweden. He is currently a Research Assistant at the Department of Economics, Stockholm University, Sweden, and a Ph.D. candidate within the Systems and Molecular Biomedicine group at the Department of Life Sciences and Medicine, University of Luxembourg. His interdisciplinary academic research spans mechanistic interpretability of deep architectures, continual fine-tuning, and multi-expert routing stability.
\endbio

\end{document}